\documentclass{article}
\usepackage{ijcai25}

\usepackage{times}
\usepackage{soul}
\usepackage{url}
\usepackage[hidelinks]{hyperref}
\usepackage[utf8]{inputenc}
\usepackage[small]{caption}
\usepackage{graphicx}
\usepackage{amsmath}
\usepackage{amsthm}
\usepackage{amssymb}
\usepackage{booktabs}
\usepackage{algorithm}
\usepackage{algorithmic}
\usepackage[switch]{lineno}
\usepackage{tcolorbox}

\usepackage{hyperref}
\usepackage{xcolor}
\definecolor{citecolor}{RGB}{0, 139, 209}
\definecolor{refcolor}{RGB}{240, 50, 50}

\definecolor{LightRed}{rgb}{0.99, 0.4, 0.2} 

\title{A Survey of Optimization Modeling Meets LLMs: Progress and Future Directions}

\author{
Ziyang Xiao$^1$
\and
Jingrong Xie$^1$\and
Lilin Xu$^1$\and
Shisi Guan$^1$\and
Jingyan Zhu$^1$\and
Xiongwei Han$^2$\and
Xiaojin Fu$^2$\and
WingYin Yu$^2$\and
Han Wu$^2$\and
Wei Shi$^2$\and
Qingcan Kang$^2$\and
Jiahui Duan$^2$\and
Tao Zhong$^2$\and
Mingxuan Yuan$^2$\and
Jia Zeng$^2$\and
Yuan Wang$^3$\and
Gang Chen$^1$\And
Dongxiang Zhang$^{1,4}$\thanks{Corresponding Author.}\\
\affiliations
$^1$The State Key Laboratory of Blockchain and Data Security, Zhejiang University\\
$^2$Huawei Noah’s Ark Lab\\
$^3$School of Business, Singapore University of Social Sciences\\
$^4$Hangzhou High-Tech Zone (Binjiang) Institute of Blockchain and Data Security\\
\emails
\{xiaoziyang, zhangdongxiang,  cg\}@zju.edu.com,
\{hanxiongwei, rocket.yuwingyin\}@huawei.com
}

\begin{document}

\maketitle

\begin{abstract}
By virtue of its great utility in solving real-world problems, optimization modeling has been widely employed for optimal decision-making across various sectors, but it requires substantial expertise from operations research professionals. With the advent of large language models (LLMs), new opportunities have emerged to automate the procedure of mathematical modeling. This survey presents a comprehensive and timely review of recent advancements that cover the entire technical stack, including data synthesis and fine-tuning for the base model,  inference frameworks, benchmark datasets, and performance evaluation. In addition, we conducted an in-depth analysis on the quality of benchmark datasets, which was found to have a surprisingly high error rate.  We cleaned the datasets and constructed a new leaderboard with fair performance evaluation in terms of base LLM model and datasets. We also build an online portal that integrates resources of cleaned datasets, code and paper repository to benefit the community.  Finally, we identify  limitations in current methodologies and outline future research opportunities.


\end{abstract}

\section{Introduction}

Optimization modeling aims to mathematically model complex decision-making problems that arise from a wide range of industry sectors, including supply chain ~\cite{DBLP:journals/jors/Cuthbertson98}, healthcare \cite{delgado2022equity}, air traffic flow management \cite{DBLP:journals/eor/MatosO00,DBLP:conf/aaai/ZhangXWSC23} and geometry problems \cite{DBLP:journals/kais/XiaoZ23}. Despite its potential to enhance operational efficiency, there exists an expertise barrier that limits the broader adoption of modeling tools. According to a survey of Gurobi users, $81\%$ of them hold advanced degrees, with nearly half specializing in operations research~\cite{gurobiReport}. 

\begin{figure}[t!]
  \centering
  \includegraphics[width=0.48\textwidth]{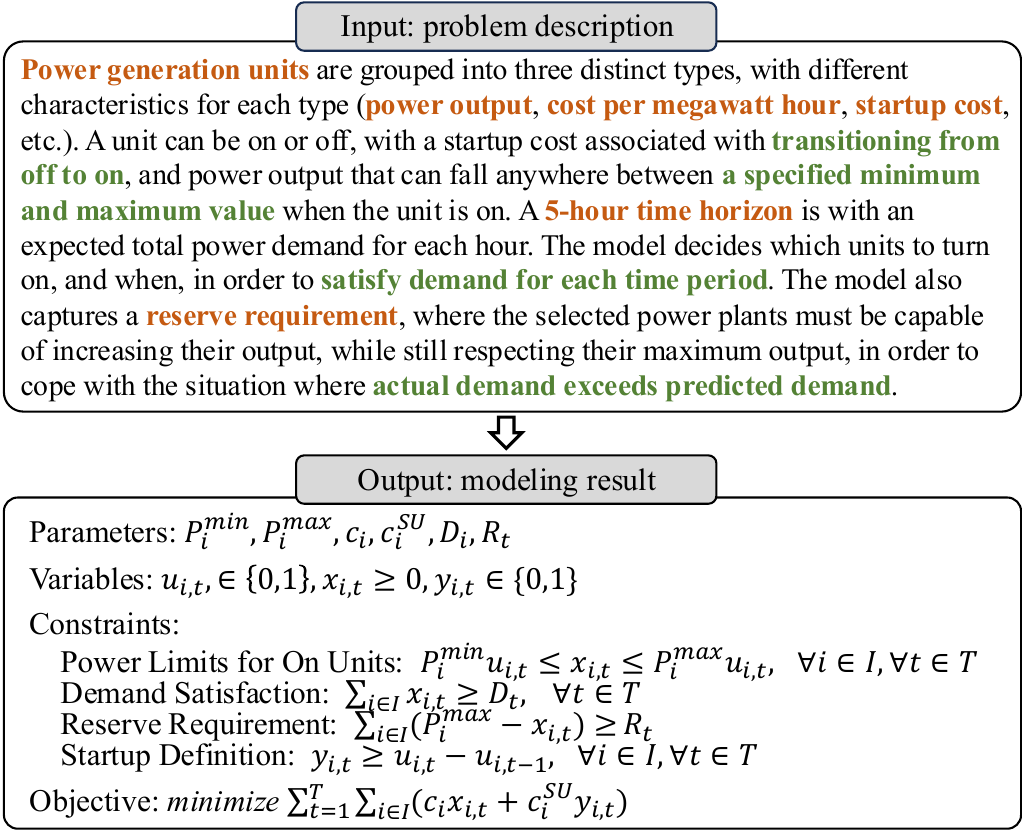}
  \caption{An example of an optimization modeling task. The orange text in the problem description implies domain-specific terminology, and the green text denotes implicit constraints.}
  \label{fig:example}
\end{figure}


To automate the procedure and reduce the dependence on domain-specific modeling experts, NL4Opt (Natural Language for Optimization) \cite{DBLP:journals/corr/abs-2303-08233} has emerged as an attractive but challenging NLP task. Its objective is to translate the text description of an OR problem into math formulations for optimization solvers. Figure~\ref{fig:example} illustrates an instance of NL4Opt task. It transforms an input problem text into a formal mathematical model, including variables, constraints, and objective function. The problem is challenging because the optimization problems often require a large amount of domain-specific knowledge to understand terminologies, such as ``megawatt hour'' and ``startup cost'', highlighted in orange text. Moreover, these descriptions may contain numerous implicit constraints that need to be inferred by human experts. Automatic optimization modeling can enhance time and cost efficiency while enabling access for users without deep optimization expertise. 


Recently, large language models (LLMs) offer a promising way to make optimization more accessible. They can understand the complicated text descriptions --- identity the optimization objective and extract the decision variables and constraints. Consequently, they automatically build the mathematical model and generate the code. 
Numerous works have been proposed in this rapidly expanding field:
\begin{itemize}
    \item \textbf{Domain-specific LLM}. Representative works such as ORLM \cite{DBLP:journals/corr/abs-2405-17743} and LLMOPT \cite{DBLP:journals/corr/abs-2410-13213} take advantage of data synthesis and instruction tuning to enhance the capability of base model for optimization modeling.
    \item \textbf{Advanced Inference Framework}: Various reasoning frameworks have emerged, include multi-agent systems (e.g. Chain-of-Experts \cite{DBLP:conf/iclr/XiaoZWXWHFZZS024} and OptiMUS \cite{DBLP:conf/icml/AhmadiTeshniziG24}) and chain-of-thought variants (e.g. Tree of Thoughts \cite{DBLP:conf/nips/YaoYZS00N23}, Autoformulation \cite{astorga2024autoformulationmathematicaloptimizationmodels}).
    \item \textbf{Benchmark Datasets and Evaluation.} There have been multiple benchmark datasets released, such as IndustryOR \cite{DBLP:journals/corr/abs-2405-17743}, NL4Opt \cite{DBLP:journals/corr/abs-2303-08233} and MAMO \cite{huang2024mamo}. However, these datasets vary significantly in quality, and evaluation methods lack standardization across different studies.
\end{itemize}

Thus, it is of high necessity to present a just-in-time survey to summarize the progress and indicate possible future research directions. In this paper, we propose the first systematic review of optimization modeling in the era of LLMs. 
As shown in Figure~\ref{fig:taxonomy}, we present a detailed taxonomy of the various methodologies employed to harness
the power of LLMs for optimization modeling. Besides, we noticed that existing benchmark datasets are associated with high error rates and performed data cleaning to enhance quality. We constructed a new leader-board with fair comparison in terms of base model and benchmark datasets, and deliver new insights of performance evaluation. To benefit the community, these datasets and implementation code are accessible from our online portal\footnote{\url{https://llm4or.github.io/LLM4OR}}. 

\section{Background}
\label{sec:background}



\subsection{Problem Definition}
Optimization modeling transforms a problem description in natural language $\mathcal{P}$ into a model $\mathcal{M}$. Mathematically, an optimization model is defined by an objective and a set of constraints, as shown in Equation \ref{eq:MILP-definition}.
\begin{align}
\begin{split}
\mathop{\text{minimize}}_{\mathbf{x}} & \quad  f(\mathbf{x}) \\ 
\text{subject to} & \quad g_{i}(\mathbf{x}) \leq 0, \quad i = 1,..., m \\ 
& \quad h_{j}(\mathbf{x}) = 0, \quad j = 1,..., p 
\end{split} \label{eq:MILP-definition}
\end{align}
Here, $\mathbf{x}$ is the vector of decision variables, $f(\mathbf{x})$ denotes the objective function, $g_{i}(\mathbf{x})$ and $h_{j}(\mathbf{x})$ represent the inequality and equality constraints respectively.

\begin{figure}[t!]
  \centering
  \includegraphics[width=0.48\textwidth]{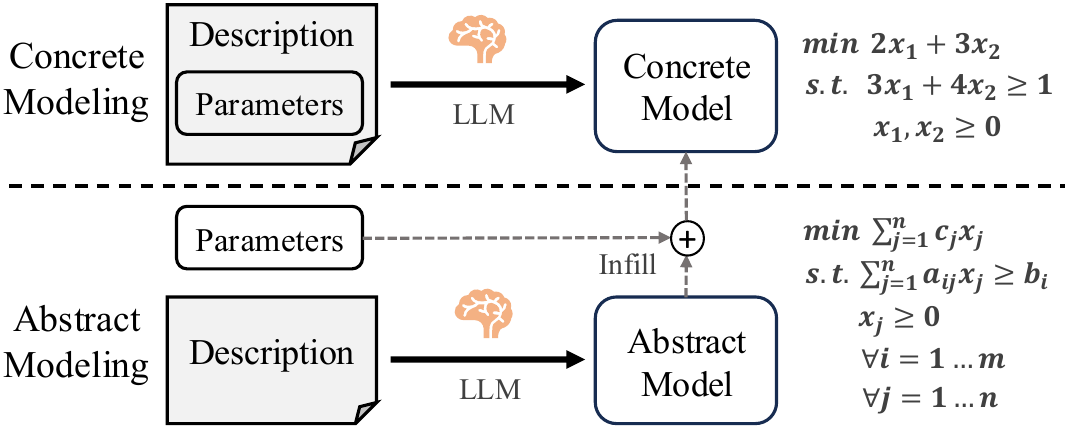}
  \caption{Comparison between concrete and abstract models. The right part illustrates a linear programming formulation example.}
  \label{fig:concrete_vs_abstract}
\end{figure}

\subsection{Abstract Model and Concrete Model}

In practice, optimization models can be categorized into two types: \emph{abstract models} and \emph{concrete models}. A model whose parameters are denoted by mathematical symbols called a  abstract model, while a model whose parameters are specified by numerical values is called a concrete model. Correspondingly, optimization modeling can be divided into two types: \emph{concrete modeling} and \emph{abstract modeling}, as illustrated in Figure~\ref{fig:concrete_vs_abstract}. Concrete modeling directly translates a problem description containing numerical parameters into a concrete model. In contrast, abstract modeling follows a model-data separation approach where the problem description only contains the model structure, with parameters are provided separately at a later stage.







\begin{figure*}[t!]
  \centering
  \includegraphics[width=1.0\textwidth]{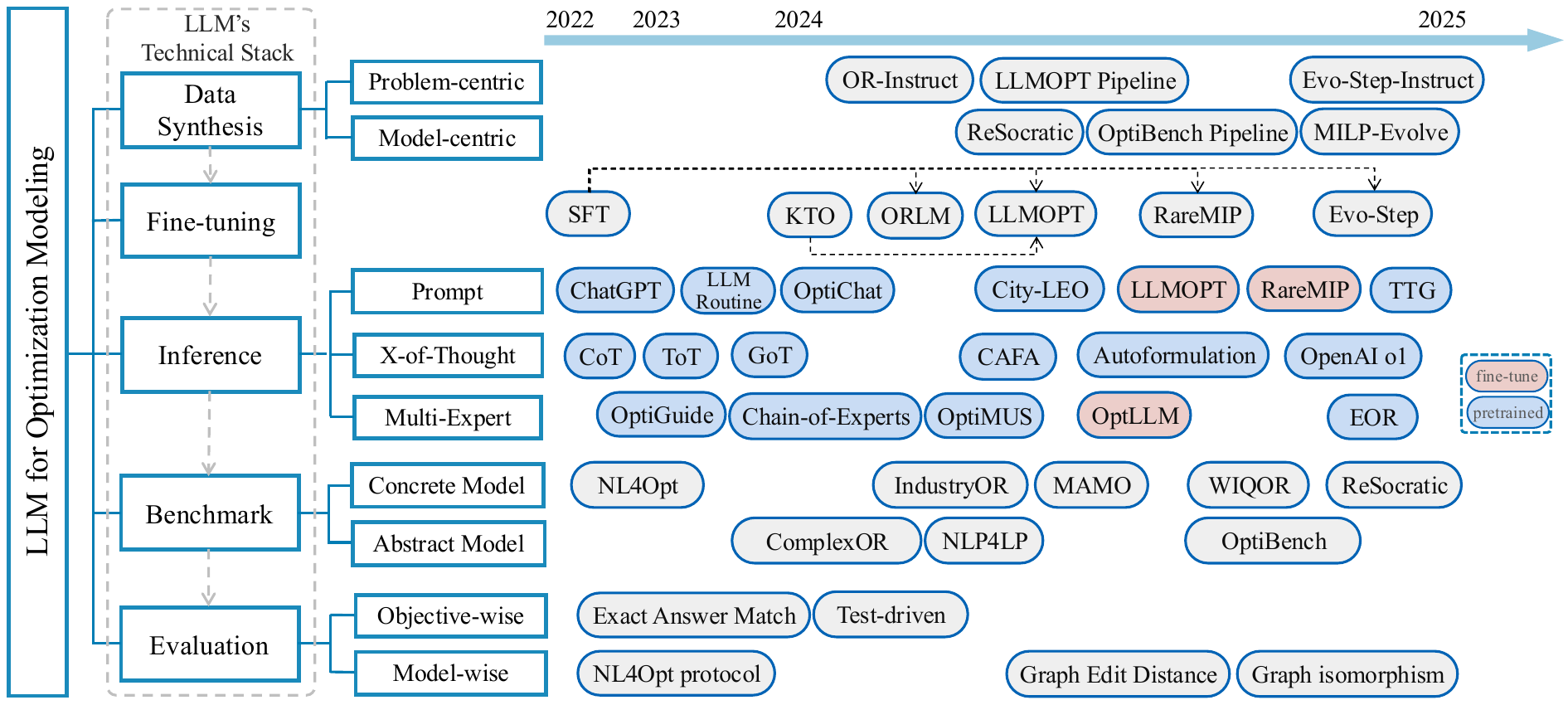}
  \caption{Left: Taxonomy of LLMs-based optimization modeling, organized according to the LLMs' technical stack. Right: Representative works for each category are presented in chronological order. The dashed arrows indicates where later works build upon techniques proposed in earlier studies.}
  \label{fig:taxonomy}
\end{figure*}

\section{Technical Stack of Optimization Modeling}

This section presents a typical technical stack for applying LLMs to optimization modeling. The pipeline consists of four key steps: (1) data preparation and LLM fine-tuning; (2) inference; (3) benchmarking; and (4) evaluation. Figure \ref{fig:taxonomy} shows the representative works in each step of this pipeline.

\subsection{Data Synthesis and Fine-tuning}


\subsubsection{Data Synthesis Methods}

It is a common practice to fine-tune language models for specialized domains such as optimization modeling. However, fine-tuning requires a substantial amount of high-quality training data. In the field of optimization modeling, data availability is limited due to the scarcity of problem sources and the high cost of problem annotation. To address this challenge, current approaches employ data synthesis to generate training datasets. Formally, the data synthesis process can be defined as $seed \rightarrow \{ \mathcal{P}', \mathcal{M}' \}$, where $\mathcal{P}'$ represents the generated problem description, $\mathcal{M}'$ denotes the corresponding modeling and $seed$ is the seed data of generation process. Depending on the primary focus of the generation process, existing works can be divided into two approaches: problem-centric and model-centric.

\paragraph{Problem-centric}


The problem-centric approach involves two steps. First, it takes an existing problem $\mathcal{P}$ and generates a new problem $\mathcal{P}'$. Then, it automatically produces the corresponding model $\mathcal{M}'$ using LLMs, with human experts filtering out low-quality annotations. In the first step, OR-Instruct \cite{DBLP:journals/corr/abs-2405-17743} devises three primitives to increase the diversity of a problem: modifying constraints and objectives, rephrasing questions for scenario diversity, and adding multiple modeling techniques for linguistic diversity. Besides, the data augmentation pipeline introduced in LLMOPT \cite{DBLP:journals/corr/abs-2410-13213} proposes seven primitives to further enhance diversity by incorporating new instructions on modifying the problem type and scenario. Beyond diversity, Evo-Step-Instruct \cite{wu2025evostep} introduces complexity as an additional dimension, along with a method to modify constraints, parameters, and objectives progressively to create more challenging problems. However, the problem-centric approach is limited in its ability to escalate complexity. As the complexity grows, generating a valid solution model becomes more difficult, leading to a higher risk of errors in annotations. To address this, Evo-Step-Instruct employs a sophisticated workflow to filter out unqualified data.

\paragraph{Model-centric}


The model-centric method adopts a different approach by first generating an augmented model $\mathcal{M}'$ and then crafting a corresponding problem description $\mathcal{P}'$. Compared to problem-centric approach, this methodology provides more fine-grained control over instance types and difficulty while ensuring the labeled model remains solvable. MILP-Evolve \cite{DBLP:journals/corr/abs-2410-08288} pioneer this approach by using existing model code as input, prompting LLMs to add, delete, or mutate code elements to evolve new models. However, since this work focus solely on generating MILP instances, it does not incorporate the problem description generation step. Similarly, OptiBench \cite{wang2024optibench} prioritizes model code generation but differs by using simple seeds such as model types (e.g., MILPs or MIPs), problem classes (e.g., knapsack problem), and domains (e.g., cargo loading) instead of existing models. This approach enables better control over dataset distribution. After code generation, LLMs transform the solver code into detailed word description. Another work, ReSocratic \cite{yang2025optibench}, extends this paradigm by defining models as semantically rich formatted demonstrations. Unlike pure code, these demonstrations incorporate structured data for variables, objective functions, and constraints, along with their natural language descriptions, resulting in richer semantic content. ReSocratic employs a multi-step sampling method with LLMs to first generate such documentation, which is then transforms into comprehensive problem descriptions as data points.


\subsubsection{Fine-tuning Methods}

Once the data is prepared, the next step is to fine-tune open-source LLMs to enhance their optimization modeling capabilities. Fine-tuning typically involves two key steps: model instruction training and model alignment. Existing works \cite{DBLP:journals/corr/abs-2405-17743,DBLP:journals/corr/abs-2409-04464,wu2025evostep} focus on the first step by applying supervised fine-tuning (SFT) with synthetic data. Meanwhile, LLMOPT \cite{DBLP:journals/corr/abs-2410-13213} introduces Kahneman-Tversky Optimization (KTO) \cite{DBLP:journals/corr/abs-2402-01306}, which further aligns model outputs with human preferences and helps mitigate biases. Despite these advancements, there remains a notable gap in research exploring innovative training techniques and paradigms for optimization modeling, highlighting the need for further investigation.

\subsection{Inference}
\label{sec:inference}

\begin{figure}[t!]
  \centering
  \includegraphics[width=0.48\textwidth]{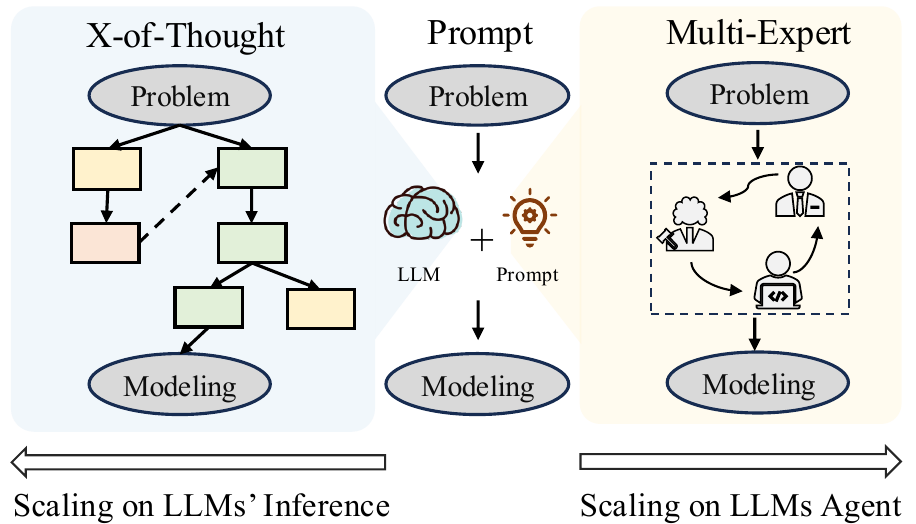}
  \caption{Three types of inference methods.}
  \label{fig:inference}
\end{figure}

During the inference stage, trained LLMs translate the problem description $\mathcal{P}$ into the modeling result $\mathcal{M}$, which can be either executable code or structured documentation. As with other domain-specific tasks, prompt engineering is a straightforward yet effective method for applying LLMs to optimization modeling problems. Moreover, as illustrated in Figure \ref{fig:inference}, the capabilities of LLMs can be enhanced along two dimensions. One approach involves inference-time scaling, which encourages LLMs to generate additional intermediate reasoning steps (referred to as ``X-of-thought''). The other approach scales up the single LLM to the LLM-based multi-agent system (referred to as ``multi-expert'').

\paragraph{Prompt}

At the advent of ChatGPT, NL4Opt \cite{DBLP:journals/corr/abs-2303-08233} pioneers the use of ChatGPT for solving optimization modeling problems. This work introduces a simple prompt template comprising three components: the problem description, task instructions, and format control. Since then, many studies have leveraged LLMs for optimization modeling via prompt engineering, which varies between training-based and training-free approaches.

For training-based approaches, prompts are primarily designed for format control, helping the model generate output that conforms to the training set's label format. For example, ORLM \cite{DBLP:journals/corr/abs-2405-17743} prompts the model to first produce a plain-text description of the model and then generate the corresponding code. Similarly, LLMOPT \cite{DBLP:journals/corr/abs-2410-13213} instructs the trained LLM to output a five-element formulation, while RareMIP \cite{DBLP:journals/corr/abs-2409-04464} prompts the model to generate LaTeX code that details the model-building process. Additionally, TTG \cite{DBLP:journals/corr/abs-2410-16456} uses prompts to produce JSON output, which can be easily parsed into a symbolic model suitable for solvers.

For training-free approaches, the goal shifts toward infusing richer domain knowledge into LLMs through the prompt. For instance, OptiChat \cite{DBLP:journals/corr/abs-2308-12923} provides the LLM with step-by-step instructions that mimic the guidance of an optimization expert, thereby equipping the model with domain-specific insights. It also employs few-shot learning by supplying examples of optimization problems paired with expert solutions. Similarly, City-LEO \cite{DBLP:journals/corr/abs-2406-10958} adopts in-context learning techniques to construct its LLM pipeline, and another work \cite{li2023llmbasedframeworkspowerengineering} incorporates prior knowledge into prompt design to further enhance LLM performance on routine tasks.

Although prompt engineering can be rapidly implemented, it only scratches the surface of what LLMs can achieve in tackling complex modeling problems. Much of their potential remains untapped. The following sections introduce two promising directions to unleash this power: X-of-thought and Multi-Agent.

\paragraph{X-of-Thought}

To enhance the reasoning capabilities of LLMs and tackle increasingly complex optimization modeling problems, researchers have begun exploring LLMs' potential during inference time. The chain-of-thought (CoT) approach \cite{DBLP:conf/nips/Wei0SBIXCLZ22} pioneers LLM reasoning by encouraging the model to think step-by-step, effectively bridging logical gaps during inference. Building on this foundation, Tree of Thoughts (ToT) \cite{DBLP:conf/nips/YaoYZS00N23} and Graph of Thoughts (GoT) \cite{besta2024got} further enhance reasoning by employing tree- and graph-structured exploration of intermediate thoughts. Collectively, these approaches are known as ``X-of-thought'' \cite{DBLP:conf/acl/ChuCCYH0P00L24}. Although originally designed for general reasoning tasks, these methods have also been successfully applied to optimization modeling \cite{DBLP:conf/iclr/XiaoZWXWHFZZS024}.

Subsequently, several X-of-thought methods tailored for optimization modeling have emerged. For instance, CAFA \cite{deng2024cafa} defines the inference process as a linear sequence of steps that explicitly captures the reasoning required for modeling. Furthermore, Autoformulation \cite{astorga2024autoformulationmathematicaloptimizationmodels} treats the modeling process as a Monte Carlo Tree Search, where each level of the tree corresponds to a specific modeling step---sequentially addressing parameters and decision variables, the objective function, equality constraints, and inequality constraints. This framework integrates an LLM with two key components: (1) a dynamic formulation hypothesis generator responsible for exploring the Monte Carlo Tree, and (2) an evaluator that provides feedback on the correctness of solutions at the leaf nodes.

Recently, OpenAI's o1 \cite{o12024} has attracted significant attention for its exceptional reasoning capabilities in tackling complex problems, including optimization modeling. It explicitly integrates an extended internal chain-of-thought into its inference process, representing a promising direction that merits further investigation.

\paragraph{Multi-Expert}

Another approach to scaling language models for complex reasoning is the use of multi-agent collaboration systems \cite{qian2024scalinglargelanguagemodelbasedmultiagentcollaboration}. In the field of optimization modeling, LLMs are adapted to mimic human experts and collaborate to complete the entire modeling process. This system is referred to as multi-expert system. Early examples include OptiMUS \cite{DBLP:conf/icml/AhmadiTeshniziG24} and Chain-of-Experts (CoE) \cite{DBLP:conf/iclr/XiaoZWXWHFZZS024}. Both systems predefine a set of LLM-based experts, with two key roles: a formulator for optimization modeling and a programmer for code generation. They differ in how they manage the workflow: OptiMUS uses a predefined workflow to engage experts in collaborative problem-solving, while CoE employs a special expert called the ``Conductor'' to orchestrate the entire process. Additionally, CoE introduces a system-level reflection mechanism to adjust answers based on external feedback.

Subsequently, the OptiGuide framework \cite{DBLP:journals/corr/abs-2307-03875} is proposed with a focus on improving the reliability and readability of modeling results. Specifically, it incorporates a safeguard agent to address potential output errors and an interpreter that generates human-readable explanations of both the modeling results and the solver's solution. Similarly, OptLLM \cite{DBLP:conf/naacl/ZhangWGWLYY24} includes a diagnostic agent that reformulates the modeling output based on internal feedback when code fails syntax tests. Explainable Operations Research (EOR) \cite{zhang2025decision} adopts a similar framework to OptiGuide but focuses on what-if analysis for optimization modeling, in which way it can evaluate the impact of complex constraint changes on decision-making.

Compared to X-of-Thought, the merits of multi-expert methods lie in their interpretable intermediate results and better capability of safeguarding against potential errors hidden in the output, making them a popular direction for future research.

\subsection{Benchmarks}


To evaluate performance of LLMs-based optimization modeling methods, several benchmarks have been proposed. As discussed in Section \ref{sec:background}, these benchmarks can be categorized into two types: concrete modeling and abstract modeling.

\paragraph{Concrete Modeling}
NL4Opt \cite{DBLP:journals/corr/abs-2303-08233} is the first optimization modeling benchmark proposed in a competition, featuring a test set of $289$ instances. However, NL4Opt primarily focuses on simple optimization modeling problems. To address the need for more challenging cases, IndustryOR \cite{DBLP:journals/corr/abs-2405-17743} is introduced, consisting of $100$ real-world industry cases. IndustryOR covers a variety of problem types—including mixed integer programming and nonlinear integer programming—and features descriptions with or without tabular data, thereby increasing problem complexity. However, IndustryOR suffers from quality control issues, which result in a high error rate. To overcome this limitation, ReSocratic \cite{yang2025optibench} introduces a comprehensive framework that applies multiple filters to remove erroneous cases, efficiently improving dataset quality and expanding the test set to $605$ instances. While the annotations in these three benchmarks focus solely on providing an objective as final answer, MAMO \cite{huang2024mamo} goes a step further by including optimal variable information, offering additional perspectives for evaluating model correctness. Note that MAMO also categorize problems into three classes: EasyLP, ComplexLP and ODE. Our study primarily focuses on the former two categories. All these benchmarks are designed for end-to-end modeling tasks. WIQOR \cite{parashar2025wiqor}, on the other hand, employs what-if analyses to assess performance, providing insights into whether LLMs possess a deeper understanding of the modeling process.

\paragraph{Abstract Modeling}
ComplexOR \cite{DBLP:conf/iclr/XiaoZWXWHFZZS024} is an abstract modeling benchmark introduced in the CoE, containing $37$ instances collected from both industrial and academic scenarios. In ComplexOR, numerical parameter values are separated from the problem descriptions. NLP4LP \cite{DBLP:conf/icml/AhmadiTeshniziG24} is another early abstract modeling benchmark, extending the number of instances to $269$. Although both datasets are relatively small, the subsequent release of OptiBench \cite{wang2024optibench} offers a larger collection of $816$ instances following a model-data separation format. While most existing research focuses on concrete modeling, abstract modeling is more prevalent in industrial scenarios due to its reusability. However, developing high-quality benchmarks for abstract modeling remains challenging because of its inherent complexity.

\subsubsection{Analysis on Benchmarks}


\begin{table}[t!]
\centering
\begin{tabular}{lrrr}
    \toprule
     \bf Dataset  & \bf Size & \bf Complexity &  \bf Error Rate \\
    \midrule 
  NL4Opt     & $289$  & $5.59$    &  $\geq 26.4\%$  \\
IndustryOR & $100$  & $\bf 14.06$    &  $\geq \bf 54.0\%$  \\
EasyLP       & $652$    & $7.12$  &  $\geq 8.13\%$  \\
ComplexLP       & $211$    & $\bf 13.35$  &  $\geq 23.7\%$  \\
ReSocratic & $605$    & $7.45$   &  $\geq 16.0\%$ \\
    NLP4LP     & $269$    & $5.58$   &  $\geq 21.7\%$  \\
  ComplexOR  & $37$     & $5.98$    &  $\geq 24.3\%$ \\
    \bottomrule
\end{tabular}
\caption{Quality statistics of optimization modeling benchmarks.}
\label{tab:benchmarks}
\end{table}

\begin{figure}[t!]
  \centering
  \includegraphics[width=0.48\textwidth]{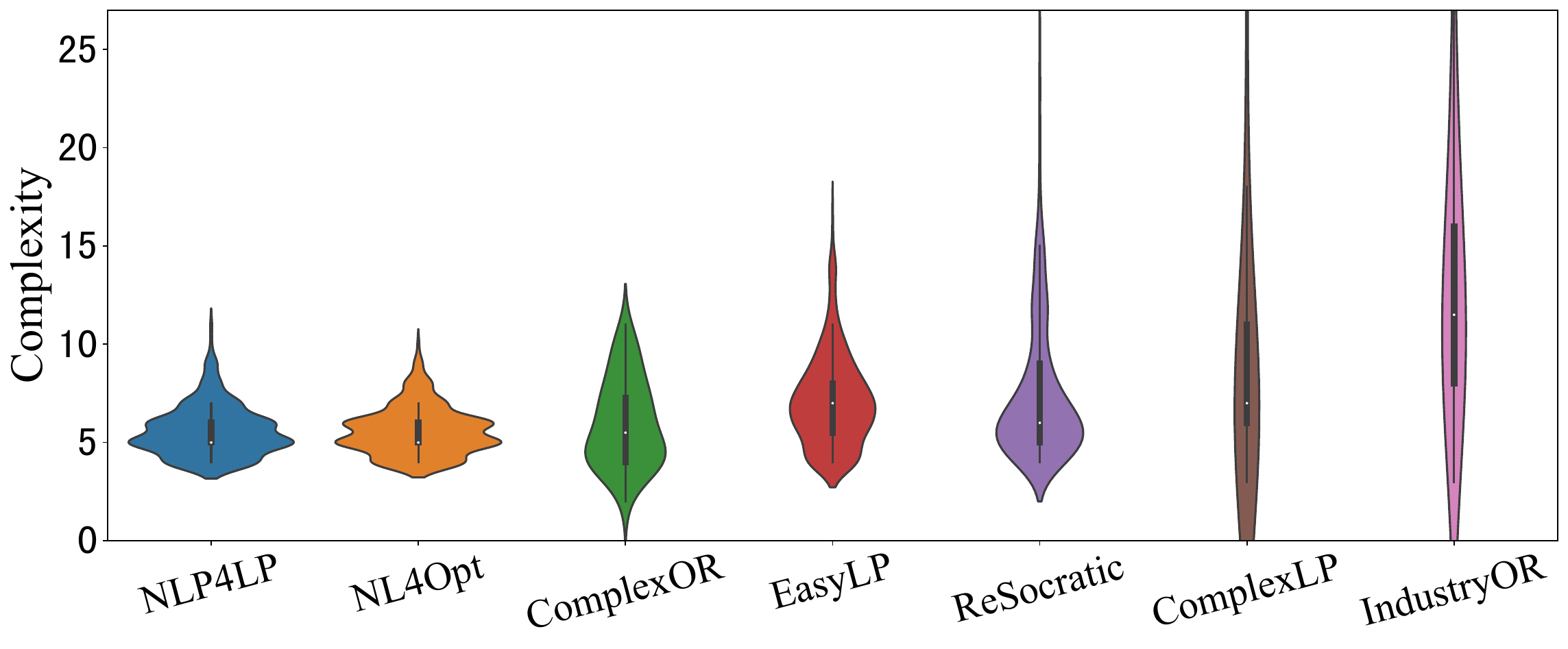}
  \caption{Statistics of complexity distribution for each benchmark visualized using a violin plot. X-axis shows different benchmarks, and y-axis shows the complexity indicator.}
  \label{fig:complexity}
\end{figure}

To assess the quality of current benchmarks, we conduct an in-depth analysis of them. The results are shown in Table \ref{tab:benchmarks} and Figure \ref{fig:complexity}. We evaluate three key statistical features: (1) Data Size: the number of instances in the benchmark's test set; (2) Complexity: for each problem, we first use standard prompting to generate a model and then use the number of variables and constraints in the model to indicate its complexity; (3) Error Rate: to compute this metric, we have $11$ human experts manually identify errors in the problems, and each error case is cross-validated by at least three different experts.


\begin{table*}[t!]
\centering
\begin{tabular}{l|lllllll}
    \toprule
     \bf Methods & \bf NL4Opt  & \bf IndustryOR & \bf EasyLP & \bf ComplexLP & \bf NLP4LP & \bf ReSocratic & \bf ComplexOR  \\
    \midrule
    Standard         &  $61.2\%$   &  $38.1\%$  & $70.3\%$ &  $57.7\%$  &  $73.6\%$ & $48.4\%$ &  $42.9\%$ \\
    CoT              &  $62.2\%$   &  $40.5\%$  &  $49.5\%$  &  $42.3\%$ & $74.7\%$  &  $43.6\%$ & $39.2\%$ \\
    Chain-of-Experts &  $66.7\%$   &  $31.2\%$  & $\bf 94.4\%$ & $50.6\%$  &  $\bf 87.4\%$ & $\bf 71.2\%$ & $\bf 57.1\%$  \\
    CAFA             &  $68.1\%$   &  $41.1\%$  &     $71.2\%$       & $44.5\%$ &    $50.0\%$    &  $40.1\%$   &  $46.4\%$ \\
    \midrule
    ORLM-\scriptsize{LLaMA-3 8B}    &  $\bf 73.8\%$   &  $\bf 42.9\%$  &  $90.4\%$  & $\bf 59.5\%$ &   $76.4\%$     & $61.8\%$  &  $50.0\%$ \\
    \bottomrule
\end{tabular}
\caption{Performance comparison of existing fully open-source methods on cleaned benchmarks in a unified setting (use GPT-4o for training-free methods and use accuracy as metric). All results are reproduced using our standardized evaluation method.}
\label{tab:evaluation}
\end{table*}

According to our results, we obtain several key findings. First, in current benchmarks, the error rate is relatively high. As shown in Table \ref{tab:benchmarks}, except for EasyLP in MAMO, the error rates of other benchmarks exceed $15\%$, with IndustryOR even reaching as high as $54\%$, indicating that these benchmarks are not entirely reliable for evaluation. The errors can be caused by three main factors: (1) logical errors in problem descriptions, such as unbounded constraints; (2) poorly defined parameters that lead to unsolvable models; and (3) incorrect ground truth data. To address these issues, we manually filter all error cases and compile a unified, cleaned collection of optimization modeling benchmarks to facilitate future research.

\begin{tcolorbox}[
    colframe=black,
    colback=white,
    arc=0pt,
    boxsep=2pt,
    left=4pt,
    right=4pt,
    top=4pt,
    bottom=4pt
]
\textbf{Takeaway \#1:} The high error rates in current benchmarks undermine their reliability. We curate a cleaned and unified set of optimization modeling benchmarks to facilitate more accurate evaluation.
\end{tcolorbox}

Second, our analysis of benchmark complexity reveals that current benchmarks mainly cover simple cases and exhibit an imbalanced distribution. As shown in Table \ref{tab:benchmarks}, NL4Opt, NLP4LP, and ComplexOR clearly present low levels of challenge. Figure \ref{fig:complexity} further shows that most instances concentrate at the simple and medium complexity levels, with instances of complexity greater than $10$ being very scarce, which indicates a lack of truly complex cases.

\begin{tcolorbox}[
    colframe=black,
    colback=white,
    arc=0pt,
    boxsep=2pt,
    left=4pt,
    right=4pt,
    top=4pt,
    bottom=4pt
]
\textbf{Takeaway \#2:} Existing benchmarks are dominated by simple and moderate problems, with very few challenging cases. This imbalance highlights the need for more high-complexity benchmarks.
\end{tcolorbox}

\subsection{Evaluation}

Evaluating optimization models can be challenging because it is often difficult to determine the correctness of the results. There are two primary approaches exist. The first is objective-wise evaluation, which focuses exclusively on the final objective value produced by the solver. The second is model-wise evaluation, where the generated model is directly compared against a ground truth model.


\paragraph{Objective-wise}

In objective-wise evaluation, the focus is solely on the correctness of the final objective. This approach originates from mathematical word problems \cite{DBLP:journals/corr/abs-2110-14168}, where LLMs directly generate a final answer and compare it to the ground truth, referred as the exact answer match method. However, in optimization modeling, LLMs produce a model rather than a final answer. To address this, a test-driven method is introduced in Chain-of-Experts (CoE) \cite{DBLP:conf/iclr/XiaoZWXWHFZZS024}, where a solver takes the generated model (with specified parameters), computes the final objective, and compares it to the ground truth. Subsequent works, including ORLM \cite{DBLP:journals/corr/abs-2405-17743}, CAFA \cite{deng2024cafa}, and Autoformulation \cite{astorga2024autoformulationmathematicaloptimizationmodels}, adopt this same test-driven method.

\paragraph{Model-wise}

While objective-wise evaluation is straightforward, it has a notable limitation: a correct objective value does not necessarily guarantee a correct model. To address this, model-wise evaluation is introduced. NL4Opt \cite{DBLP:journals/corr/abs-2303-08233} pioneers a protocol that converts modeling results into a canonical formulation, where the coefficients of the objective function and constraints are extracted into matrices and then are compared with ground truth. Although this method captures model correctness comprehensively, it provides only a binary metric and fails to reflect the degree of correctness, which is essential for fine-grained assessments. To overcome this limitation, a graph-based evaluation method \cite{DBLP:conf/coling/XingWFFXGFRMHZ024} is proposed, representing modeling results as a graph and using graph edit distance to produce a continuous correctness score between $0$ to $1$. Building on this, a modified graph isomorphism testing algorithm \cite{wang2024optibench} offers even more precise evaluation, with theoretical guarantees ensuring the correctness of its comparisons.

\subsubsection{Evaluation Result of Existing Methods}

In this survey, we observe that the reported evaluation results across existing works often exhibit inconsistencies, making fair comparisons challenging. These discrepancies arise primarily from three factors.

\begin{itemize}
    \item \textbf{Choice of Base Model}: Researchers use different commercial LLMs as base model. For example, Chain-of-Experts employs GPT-3.5, whereas Autoformulation uses GPT4-mini, due to the rapid evolution of LLMs.
    
    
    \item \textbf{Dataset Preprocessing Approaches}: Different strategies are used for handling incorrect samples and decimal precision, resulting in varying preprocessing pipelines.
    
    \item \textbf{Evaluation Metrics}: Metrics also vary: ORLM reports micro and macro average accuracy, whereas Chain-of-Experts focuses on compile error rates.
\end{itemize}

These factors collectively contribute to the difficulty of establishing a consistent leader-board for optimization modeling methods.

To address the challenge of inconsistent evaluations and create a fair comparison, we adopt a unified setting to assess all fully open-source optimization modeling methods on our cleaned benchmarks. Specifically, we employ the cutting-edge commercial LLM \textit{gpt-4o-2024-08-06} as the base model for all training-free methods. We report accuracy as the evaluation metric, as it is the most widely accepted measure.

Regarding optimization modeling methods, we strive to evaluate every fully open-source approach. However, many methods mentioned in Subsection \ref{sec:inference} remain closed-source, including LLMOPT \cite{DBLP:journals/corr/abs-2410-13213}, RareMIP \cite{DBLP:journals/corr/abs-2409-04464}, Autoformulation \cite{astorga2024autoformulationmathematicaloptimizationmodels}, OptLLM \cite{DBLP:conf/naacl/ZhangWGWLYY24}, LLM Routine \cite{li2023llmbasedframeworkspowerengineering}, City-LEO \cite{DBLP:journals/corr/abs-2406-10958}, and TTG \cite{DBLP:journals/corr/abs-2410-16456}. Three other methods, including OptiChat~\cite{DBLP:journals/corr/abs-2308-12923}, OptiGuide~\cite{DBLP:journals/corr/abs-2307-03875}, and EOR~\cite{zhang2025decision}, are interactive and thus not directly comparable to end-to-end approaches. Additionally, OptiMUS~\cite{DBLP:conf/icml/AhmadiTeshniziG24} requires a preprocessing step that is unavailable for most benchmarks, leading us to exclude it. For broader comparison, we include standard and chain-of-thought prompting as baselines.

\begin{tcolorbox}[
    colframe=black,
    colback=white,
    arc=0pt,
    boxsep=2pt,
    left=4pt,
    right=4pt,
    top=4pt,
    bottom=4pt
]
\textbf{Takeaway \#3:} The evaluation results reported in existing works lack a unified standard. And the open-source landscape in optimization modeling remains limited.
\end{tcolorbox}

Table \ref{tab:evaluation} shows the overall results, revealing several key observations. First, Chain-of-Experts and ORLM are two competitive methods in optimization modeling. While Chain-of-Experts works well for simpler tasks, ORLM surpasses it on more complex datasets such as IndustryOR and ComplexLP, indicating that trained models may be more effective in challenging scenarios. Second, contrary to popular belief, CoT does not always yield better results than standard prompting. On certain datasets, it even leads to a noticeable drop in performance, supporting the idea that CoT should be applied selectively \cite{sprague2024cotcotchainofthoughthelps}. Finally, the performance of CAFA is comparable to CoT, likely as CAFA can be viewed as a specialized form of CoT prompting. 

\begin{tcolorbox}[
    colframe=black,
    colback=white,
    arc=0pt,
    boxsep=2pt,
    left=4pt,
    right=4pt,
    top=4pt,
    bottom=4pt
]
\textbf{Takeaway \#4:} Three key findings: (1) Chain-of-Experts and ORLM emerge as the most competetive frameworks; (2) CoT prompting does not always outperform standard prompting; (3) The performance of CAFA resembles that of a specialized CoT strategy.
\end{tcolorbox}

\section{Online Portal for Optimization Modeling}
We develop a website portal that integrates the resources of LLM-based optimization modeling and provides great convenience for researchers to follow the topic. First, we provide the download links for both original  and cleaned version of benchmark datasets. Second, we collect and publish the implementation of existing solutions and provide a leader-board to report their performance on the benchmarks. Thirdly, we continue to update the latest research papers on this promising research domain. We believe such an integrated portal brings significant benefit for the community.

\section{Challenges and Future Directions}





\subsection{Reasoning Model for Optimization Modeling}

A prominent trend in recent LLM research is enhancing the reasoning capabilities of base models. The release of OpenAI o1 \cite{o12024} demonstrates impressive performance on complex mathematical tasks. However, these advances have not yet been transferred to optimization modeling. One key obstacle is that training a reasoning model heavily relies on long chain-of-thought data, which is difficult to annotate in the context of optimization modeling. To bridge this gap, Deepseek R1 Zero \cite{DBLP:journals/corr/abs-2501-12948} proposed a promising alternative by using pure reinforcement learning for training, enabling LLMs to develop reasoning capabilities without requiring supervised chain-of-thought annotations. This reinforcement learning strategy is also promising for optimization modeling, where the modeling process can be formulated as a Markov Decision Process and solver feedback can be used as reward to train the reasoning model.

\subsection{Explainable Modeling Processes}

The black-box nature of LLMs, most existing studies treat optimization modeling as an end-to-end process. However, the explainability of this process is also crucial for real-world applications, as it allows experts to effectively debug, modify, and understand the generated models. Recent work like Explainable Operations Research \cite{zhang2025decision} has made progress in this direction by developing methods to evaluate how modeling decisions impact outcomes. More research efforts to develop a trustworthy and user-friendly modeling framework are encouraged.



\subsection{Domain Knowledge Injection}

The optimization modeling process relies heavily on domain knowledge \cite{DBLP:conf/nips/XiaoZHFYZWWYC24}. As demonstrated by a research \cite{Runnwerth_2020}, much of this specialized knowledge, including conception and empirical insights, can be stored in a knowledge graph. Incorporating such domain-specific knowledge into LLMs to aid the modeling process remains a significant challenge. A recent work \cite{zhang2024chainofknowledgeintegratingknowledgereasoning} uses rule mining to construct training data from knowledge graphs and introduces a learning method to integrate knowledge graphs with LLMs, offering a promising pathway for advancing the field of optimization modeling.

\subsection{Human-in-the-Loop Modeling}

Existing inference approaches have primarily focused on the modeling capabilities of LLMs and have not explored human intervention during the inference process. Recent research indicates that LLMs can proactively query humans for domain-specific knowledge when needed \cite{pang2024empoweringlanguagemodelsactive}. These characteristics offer an opportunity to open up a new paradigm, human-in-the-loop modeling, where human experts contribute external knowledge, clarifications, and insights at critical points. To develop such a collaborative system, we need to overcome the following challenges. First, effective mechanisms are needed to identify when human intervention is required, since LLMs themselves lack this capability. Second, an effective human-in-the-loop framework should ensure that humans can seamlessly integrate their expertise into the inference process.

\section{Conclusion}

This survey provides a timely overview of the rapid progress in applying LLMs to optimization modeling. We present a thorough taxonomy of existing works across data synthesis, model fine-tuning, inference approaches, benchmarks, and evaluation methods, offering a structured understanding of the technical stack. We also highlight persisting challenges, particularly in data quality and evaluation protocols, that hinder reliable performance comparisons. To address these gaps, we evaluate current open-source methods on a set of cleaned and standardized benchmarks, revealing several key insights. Building on these findings and the latest advances, we propose promising directions to inspire further research in this emerging field.

\section*{Acknowledgments}
The work is supported by the National Key Research and Development Project of China (2022YFF0902000), the Fundamental Research Funds for the Central Universities (226-2024-00145, 226-2024-00216).

\appendix




\bibliographystyle{named}
\bibliography{ijcai25}

\end{document}